\documentclass[lettersize,journal]{IEEEtran}
\usepackage{amsmath,amsfonts}
\usepackage[
  detect-weight,
  exponent-product=\cdot,
  group-separator=.,
]{siunitx}
\DeclareSIUnit{\pp}{\textup{p.p.}}
\usepackage{lmodern}
\usepackage{algorithmic}
\usepackage{algorithm}
\usepackage{array}
\usepackage{hyperref}

\usepackage[caption=false,font=normalsize,labelfont=sf,textfont=sf]{subfig}
\usepackage{textcomp}
\usepackage{stfloats}
\usepackage{url}
\usepackage{tikz}
\usetikzlibrary{positioning}
\usepackage{verbatim}
\usepackage{graphicx}
\usepackage{import}
\usepackage[symbol*]{footmisc}

\makeatletter

\newcommand{\Rmnum}[1]{\expandafter\@slowromancap\romannumeral #1@}
\makeatother
\usepackage{cite}
\usepackage{booktabs}
\def\eg{\emph{e.g.}}

\def\etal{\emph{et al.}}

\begin{document}

\title{DBAT: Dynamic Backward Attention Transformer for Material Segmentation with Cross-Resolution Patches}

\author{Yuwen Heng, Srinandan Dasmahapatra, Hansung Kim\\School of Electronics and Computer Science \\ University of Southampton, UK}


\IEEEpubid{This paper is under consideration at Computer Vision and Image Understanding}

\maketitle

\begin{abstract}
The objective of dense material segmentation is to identify the material categories for every image pixel. Recent studies adopt image patches to extract material features. Although the trained networks can improve the segmentation performance, their methods choose a fixed patch resolution which fails to take into account the variation in pixel area covered by each material. In this paper, we propose the Dynamic Backward Attention Transformer (DBAT) to aggregate cross-resolution features. The DBAT takes cropped image patches as input and gradually increases the patch resolution by merging adjacent patches at each transformer stage, instead of fixing the patch resolution during training. We explicitly gather the intermediate features extracted from cross-resolution patches and merge them dynamically with predicted attention masks. Experiments show that our DBAT achieves an accuracy of 86.85\%, which is the best performance among state-of-the-art real-time models. Like other successful deep learning solutions with complex architectures, the DBAT also suffers from lack of interpretability. To address this problem, this paper examines the properties that the DBAT makes use of. By analysing the cross-resolution features and the attention weights, this paper interprets how the DBAT learns from image patches. We further align features to semantic labels, performing network dissection, to infer that the proposed model can extract material-related features better than other methods.  We show that the DBAT model is more robust to network initialisation, and yields fewer variable predictions compared to other models. The project code is available at \href{https://github.com/heng-yuwen/Dynamic-Backward-Attention-Transformer}{https://github.com/heng-yuwen/Dynamic-Backward-Attention-Transformer}.

\end{abstract}

\begin{IEEEkeywords}
Material segmentation, image processing, scene understanding, neural networks, network interpretability.
\end{IEEEkeywords}

\section{Introduction}
\label{sec:intro}
The dense material segmentation task aims to recognise the physical material category (\eg{} metal, plastic, stone, etc.) of each pixel in the input image. The predicted material labels are instrumental in various applications such as autonomous robots or entertainment systems. For example, the material information is essential in robot manipulation \cite{zhao2020simultaneous,shrivatsav2019tool} and is also used for immersive sound rendering by environment analysis \cite{mcdonagh2018synthesizing,kim2019immersive,chen2020context,heng2023material,mona2022audio}. Since the  appearance such as shape, colour and transparency of a specific material can vary, identifying the materials from general RGB images remains a challenging task \cite{8675400,heng22camseg,bell15minc}. In order to improve predictive accuracy, recent material segmentation methods propose to combine both material and contextual features \cite{schwartz2018visual,8675400,schwartz2016material,bell2015material,heng22camseg}. Material characteristics such as texture and roughness allow models to identify material categories without knowing all their varied appearances. Contextual features including objects and scenes can limit the possible categories of materials in the image. 

However, existing methods have not thoroughly investigated different strategies to combine the material and contextual features. The network proposed by Schwartz \etal \cite{8675400,schwartz2016material,schwartz2018visual} extracts these features with independent branches and concatenates them to make the material decisions. The material features are extracted from cropped image patches and contextual features, including objects and scenes, are gathered from multiple pre-trained branches targeting related tasks, such as object and scene classification. In our preliminary work \cite{heng22camseg}, we extended this multi-branch architecture with a self-training strategy \cite{cheng2020weakly,zoph2020rethinking} to boundary features between adjacent material regions \cite{bokhovkin2019boundary}. We found that the material features generalise better when trained with a medium-sized dataset. However, the image patch resolution is fixed, and this may not be optimal for extracting material features due to the objects being at different distances to the camera. Ideally, small-resolution patches should be applied to separate adjacent material regions, while large-resolution patches can be used to cover as large a single piece of material as possible to provide sufficient material information.


\IEEEpubidadjcol
In our recent conference publication \cite{Heng_2022_BMVC}, we developed a cross-resolution transformer architecture called DBAT. The DBAT extracts material features from multi-resolution image patches rather than searching for an optimal patch resolution. Initially, images are propagated through a transformer backbone that processes 4x4 patches and gradually increases the patch size by merging adjacent patches. Subsequently, we introduce a Dynamic Backward Attention (DBA) module that aggregates the intermediate features extracted from cross-resolution image patches. The proposed DBAT is evaluated by comparing its performance on two material segmentation datasets, the Local Material Database (LMD) and OpenSurfaces, with real-time state-of-the-art (SOTA) networks. The experiments reveal that the DBAT achieves the highest accuracy and narrowest uncertainty bounds. Notably, when trained with learning rate warm-up \cite{gotmare2018closer} and polynomial decay \cite{mishra2019polynomial}, the average pixel accuracy (Pixel Acc) of the DBAT reaches 86.85\% when assessed on the LMD. This Pixel Acc represents a 21.21\% increase compared to the most recent publication \cite{heng22camseg} and outperforms the second-best model \footnote{Among the models that can serve real-time inference.} evaluated in the paper by 2.54\%.

However, similar to other network-based methods, the DBAT faces challenges in terms of interpretability. Ascertaining whether the network genuinely acquires material features through numerical evaluation or segmentation visualisation is a complex task. Therefore, in this extended paper, we further endeavour to interpret the network behaviour of the DBAT using statistical and visual tools, such as calculating the attention head equivalent patch size, visualising attention masks, and assessing the Centred Kernel Alignment (CKA) heatmap \cite{nguyen2020wide, raghu2021vision}. In order to interpret the features with human-readable concepts, we also employ the network dissection method \cite{zhou2018interpreting, bau2017network, bau2019gan, bau2020understanding} to identify the features learned by the network by aligning layer neurons with semantic concepts. By analysing the semantic concepts of the extracted features, this paper illustrates that the DBAT excels in extracting material-related features, such as texture, which is an essential property for distinguishing between various materials. By comparing the semantic concepts of features extracted by other networks trained with either material or object datasets, the results also indicate that the network architecture can influence the extracted features, and the patch-based design is indeed effective in compelling the networks to segment images based on material features.


\section{Background}
\subsection{Material Segmentation} 
Dense material segmentation aims to identify the material label for each pixel. To predict the labels, recent segmentation studies tend to choose the encoder-decoder architecture which down-samples the input image in the encoder to extract features, and up-samples the features to the original resolution in the decoder to make predictions \cite{xing2020encoder,liu2021swin,zoph2020rethinking,chen2018encoder,zhao2017pyramid,kim2018parallel}. As out-of-the-box segmentation architectures, these networks are proved to work well on many datasets. However, for the material segmentation task, no convincing performance has been achieved in the literature so far \cite{schwartz2016material,heng22camseg,zhao2020simultaneous}. Their reported per-pixel accuracy (Pixel Acc) is around 70\%. According to Schwartz \etal \cite{schwartz2016material,8675400}, annotating the materials in images is challenging due to the considerable variation in material appearance. This difficulty results in sparse annotations and unbalanced material labels in material datasets such as LMD \cite{schwartz2018visual,8675400}, OpenSurfaces \cite{bell13opensurfaces} and MINC \cite{bell15minc}.  This restricts the accuracy of deep neural networks, whose success relies on the existence of large labelled datasets. To achieve a satisfactory performance, Schwartz \etal \cite{8675400,schwartz2018visual} proposed to extract features from image patches with the assumption that the local material features can generalise the performance of the network to unseen material regions. We further observed that the networks composed of fast pooling and atrous convolution \cite{chen2018encoder,qiao2021detectors,panboonyuen2020semantic,xu2021acclvos} can overlook small material regions due to the absence of local features \cite{heng22camseg}. Both works adopt multi-branch networks to combine local material features with contextual features about objects, scenes and material boundaries. Although those studies improved the performance on material datasets, the patch resolution is fixed to 48-by-48. Fixing the size is unlikely to be optimal for all images as the areas that materials cover vary within and across images. Moreover, the multi-branch architecture is computationally costly. As a single-branch network, our DBAT is efficient enough to serve real-time applications. Furthermore, the DBAT learns from patches with multiple resolutions and dynamically adjusts the dependence on each resolution for each pixel of the feature map.

\subsection{Transformers in Vision Tasks} 
The transformer architecture exploits predictive correlations across multiple segments of the input data in parallel.  While these originated in capturing word correlations in natural language processing \cite{devlin2018bert,vaswani2017attention}, similarly weighted pixel correlations have been successfully deployed in vision tasks such as classification \cite{dosovitskiy2020image,chen2021crossvit,qing2021improved} and segmentation \cite{liu2021swin,strudel2021segmenter,zheng2021rethinking}. The core component in the transformer architecture is self-attention \cite{zhao2020exploring}, which predicts the attention score of each pixel against all other pixels, and calculates the output by adding the score-weighted pixel values (V). The attention score is obtained through the correlation between instance-specific queries (Q) and keys (K), which are predicted from the same set of pixel features. Based on the window size of the self-attention module, transformers can be categorised as either global or local types. The global transformers represented by ViT \cite{dosovitskiy2020image} and DeiT \cite{touvron2021training} apply the self-attention module to the whole feature map. This design ensures that the network can have a global view at shallow layers close to the input image. However, the quadratic complexity in attention window size makes global transformers expensive to use. Furthermore, the necessity of a global view at shallow layers is still under debate. A recent study \cite{raghu2021vision} shows that in models where global attention is employed, the learned correlations can still focus on local regions at shallow layers. Their work states that the information hidden in local regions is necessary for good performance. Since material features, such as texture, are usually extracted from local image regions, the local transformer architecture is chosen in this paper. A typical example is the Swin transformer \cite{liu2021swin,liu2021swin2}, which applies the self-attention module to small windowed regions. This local design ensures that the network learns material features independently of contextual information from the RGB images. By merging adjacent extracted patch features at each attention stage, the local transformer progressively increases the patch resolution during forward propagation. Consequently, the DBAT can predict material labels by aggregating these cross-resolution features, enabling it to handle materials that cover varying areas.

\subsection{Network Interpretability}
The study of network interpretability aims to explain how a network combines learned features to make predictions. For CNNs, the visualisation of the convolutional kernel weights shows the pattern of features that the network has extracted \cite{NIPS2012_c399862d,wang2020cnn}. For the attention-based network module, a simple yet effective way is to plot the per-pixel attention masks on the input image and the weights indicate the contribution to the final decision of each pixel \cite{fukui2019attention,liu2018picanet}. For transformers, however, interpreting the self-attention module remains challenging due to the high dimensionality of the correlation mask and the recursively connected attention modules. Carion \etal \cite{carion2020end} proposed to reduce the dimensionality by visualising an attention mask for individual pixels of the feature map one at a time. Chefer \etal \cite{chefer2021transformer} reassigned a trainable relevancy map to the input image and propagate it through all the self-attention layers. However, these methods are designed for classification tasks and they can only interpret the transformer behaviour for a specific image. In this paper, we focus on the segmentation task and we prefer a summary explanation of the whole dataset. Therefore, we choose to plot the CKA heatmap \cite{nguyen2020wide,raghu2021vision} and adopt the network dissection method \cite{bau2020understanding,bau2017network,bau2019gan} explained in the following sections.

\subsubsection{Centered Kernel Alignment}
The CKA matrix measures the layer similarity by normalising the Hilbert-Schmidt independence criterion \cite{song2012feature}, as shown in Equation (\ref{eq3}), where $X, Y$ are the feature maps extracted from two network layers. The HSIC$_1$ in Equation (\ref{eq4}) stands for the unbiased estimator of the Hilbert-Schmidt independence criterion \cite{song2012feature}, which measures the distribution alignment between $K, L$. Here $\tilde K \textnormal{ and } \tilde L$ are the matrix whose diagonal entries are set to zero. Since HSIC$_1 = 0 $ indicates the independence of $K$ and $L$, and CKA is robust to isotropic scaling, they together enable a meaningful comparison of two networks \cite{kornblith2019similarity,raghu2021vision}. 

\begin{equation}
    \textnormal{CKA}(X,Y) = \frac{  \textnormal{HSIC$_1$} (XX^T, YY^T)} {\sqrt{\textnormal{HSIC$_1$}(XX^T, XX^T)\textnormal{HSIC$_1$}(YY^T, YY^T)}}
\label{eq3}
\end{equation}

\begin{equation}
\begin{split}
     \textnormal{HSIC$_1$}(K, L) =  \frac{1}{m (m - 3)} \bigg[ tr (\tilde K \tilde L) + \frac{1^\top \tilde K 1 1^\top \tilde L 1}{(m-1)(m-2)} - \\ \frac{2}{m-2} 1^\top \tilde K \tilde L 1 \bigg]
\end{split}
\label{eq4}
\end{equation}
 Since the CKA does not need to know the network architecture, it can be used to compare the features that two arbitrary networks learn. By computing the CKA of the elements in the Cartesian product of the layer sets of two networks, we have the CKA matrix whose element CKA$_{ij}$ indicates the similarity between layer $i$ in network 1 and layer $j$ in network 2. In this paper, we show how the DBAT modules alter the features extracted against those from the backbone transformer. We show that the aggregated cross-resolution patch features are different from its backbone features and this indicates that our DBAT learns something new to improve performance. 

\subsubsection{Network Dissection} 

Visualisation tools and CKA help to understand how the model combines specific features to predict material labels and highlight the similarity or independence of features acquired by different network layers. However, they do not offer insight into the nature of these features. To address this issue, we utilise techniques from the \textquotesingle network dissection\textquotesingle{} literature \cite{bau2017network, zhou2018interpreting, bau2019gan, bau2020understanding}, which correlate neuron outputs to an independent set of human-interpretable labels, such as objects, textures, or scenes.

To calculate the correlation, we require a densely labelled dataset containing labels for the concepts. In this study, we use the Broden dataset proposed by \cite{bau2017network} to interpret the networks. First, the trained parameters are frozen. Then, the output of each neuron in the last network layer is thresholded into a mask to be compared with the corresponding concept labels in terms of mean Intersection over Union (mIoU) \cite{bau2017network}. The threshold is the value $a_k$ ensuring that 99.5\% of the activation values are greater than it. A neuron is assigned the interpretive label for which the mIoU score is the highest and above 0.04. By measuring the number of neurons aligned with each concept, the network dissection method indicates the features the network focuses on during training.

This paper applies the network dissection method to compare the proposed DBAT with selected networks. The results show that the DBAT is particularly good at detecting local material features, such as texture, which may be the reason why DBAT achieves the narrowest uncertainty bound across five runs.

It is worth noting that the network dissection method can only interpret disentangled neurons. This means that only a fraction of the channels of a network layer can be aligned with meaningful semantic concepts. The rest of the neurons also detect useful features, but they cannot be explained. One of the reasons is that these neurons are detecting mixed features (\eg detecting both texture and object combinations). One related research topic is \textquotesingle Interpretable Networks\textquotesingle. The idea is to disentangle the patterns that each neuron learns so that the visualisation of the feature map becomes interpretable. \cite{zhang2018interpreting} proposed to separate the patterns that a network learns by building an explanatory graph. The explanatory graph can be applied to trained networks and summarise the extracted features into a few patterns. \cite{zhang2018interpretable} further introduced a filter loss term to regularise the features so that each neuron contributes to one category with one consistent visual pattern. However, their networks can only learn features from ball-like areas since the filter loss is based on a regional template.  \cite{shen2021interpretable} extended the interpretable networks to learn disentangled patterns without shape or region limitations. Their compositional network splits neurons into groups and makes the neurons learn similar/different features within/across the groups. The trained networks can produce meaningful feature maps with a slight sacrifice in accuracy \cite{shen2021interpretable}. However, training an interpretable network is beyond the scope of this paper and we will investigate this method in the future.

\section{Dynamic Backward Attention Transformer}
 This section explains the DBAT structure in detail. As shown in Fig \ref{net_overview}, the DBAT consists of three modules: the backbone encoder, the dynamic backward attention module, and the feature merging module. The encoder is responsible for extracting cross-resolution feature maps with window-based attention and patch merging. The proposed Dynamic Backward Attention (DBA) module predicts per-pixel attention masks to aggregate the cross-resolution feature maps extracted from the encoder. The feature merging module guides the DBA module to extract features that are complementary to the last stage encoder output, which holds a global view of the image. Finally, the merged features, which have both enhanced cross-resolution features as well as global features, are passed into a segmentation decoder to make the material predictions. 


\subimport{../tikzfiles/}{netarchitecture.tex}

\subsection{Dynamic Backward Attention Module}
\label{dbasection}
The DBA module depends on a backbone encoder to extract feature maps from cross-resolution patches. There are multiple approaches to design the encoder. One possible choice is to employ multiple branches that learn from varying-sized patches, sharing the features during training \cite{heng22camseg}, and tuning the number of trainable parameters with the patch size. Another option is to utilise non-overlapping convolutional kernels \cite{yamanakkanavar2020using} and enlarge the patch size through a pooling layer.

In this paper, we find the transformer as a suitable encoder to extract cross-resolution patch features \cite{heng22camseg, 8675400}, as it is inherently designed to process image patches \cite{dosovitskiy2020image} and demonstrates promising results for vision tasks \cite{dosovitskiy2020image, liu2021swin, liu2021swin2, radford2021learning}. Another reason is that, according to recent research \cite{raghu2021vision}, the self-attention module can adapt its equivalent attention distance by assigning weights to each pixel. In this study, the equivalent attention distance is defined as the Euclidean distance between two pixels, weighted by the attention weight. If the predictive accuracy on the training set improves by increasing the attention weights of neighbouring pixels, the network is considered to prefer local features. By allowing the network to choose from a large number of patch sizes through attention weights, our DBAT encoder can efficiently encode features at different resolutions, despite undergoing only four stages, which will be discussed in Section \ref{attention_analysis}.

Fig. \ref{dbamodule} shows the way our DBA module aggregates the cross-resolution features. With the assumption that the features at each transformer stage can preserve spatial location information \cite{raghu2021vision}, we propose to aggregate these features through a weighted sum operation for each pixel of the feature map. For stage $i$, the feature map spatial size can be computed as $(\frac{H}{2\times2^i}, \frac{W}{2\times2^i})$, where $H$ and $W$ are the input image height and width. The attention weights $Attn_i$ are predicted from the last feature map, $Map_4$ with a 1$\times$1 convolutional layer. To perform the aggregation operation, it is necessary to normalise the attention weights with softmax so that the weights across the masks sum to 1 at each pixel location. It is worth noting that the spatial shapes of the feature maps should be the same in order to perform the pixel-wise product between $Map_i$ and $Attn_i$. Moreover, the shapes of $Attn_i$ should be the same as well to normalise the per-pixel attention masks. In this paper, we set $Attn_i$ to be the same size as $Map_4$, and down-sample $Map_{1,2,3}$ to the shape of $Map_4$ so that the computation and memory overhead can be minimised. The attention mechanism is expressed by Equation (\ref{eq2-2}). Here the product and sum operations are all performed element-wise.
\begin{align}
    Aggregated\ Feature &= \sum_{i=1}^{i=4}Attn_i \odot Map_i
    \label{eq2-2}
\end{align}
\subimport{../tikzfiles/}{dbatmodule.tex}     
\subsection{Feature Merging Module}
Although the DBA module aggregates cross-resolution features, it is not guaranteed that the aggregated features can improve network performance. Moreover, it is not desired to drop all global or semi-global features since they can limit the possible material categories in a given context \cite{schwartz2016material}. Therefore, the feature merging module is proposed to guide the DBA module to enhance the local features without harming the original backbone performance. This module consists of a global-to-local attention as well as a residual connection \cite{he2016deep}. The simple residual connection, \eg \  $Merged\ Feature = Map_4 + Aggregated\ Feature$, can ensure the DBA module learns complementary features. However, this simple addition operation would over-emphasise $Map_4$ and break our DBA module which aggregates features linearly. Therefore, we choose to bring in non-linear operations with the global-to-local attention \cite{shen2022effects,liu2021swin,tu2021hyperspectral,xu-etal-2019-leveraging}, which identifies the relevant information in the aggregated cross-resolution patch features through attention mechanism, and merges it into $Map_4$, as illustrated in Fig. \ref{featuremerging}. The query matrix Q is predicted from $Map_4$ and it is applied to the key matrix K from the aggregated features. The matrix multiplication (represented by the @ symbol) between Q and $\text{K}^T$ produces the attention alignment scores. Then the scores are normalised with softmax to extract relevant information from the value matrix V. Here the window-based attention in \cite{liu2021swin} is used as well. With the DBA module and the feature merging module, the mechanism of DBAT can be described as enhancing the material-relate local features by injecting cross-resolution patch features into $Map_4$.
\subimport{../tikzfiles/}{residual.tex}

\section{Experiment Setting}
\subsection{Material Segmentation Datasets} 
\label{dataset_dis}
The present study evaluates the proposed DBAT using two datasets, namely the LMD \cite{schwartz2016material,8675400,heng22camseg} and the OpenSurfaces \cite{bell13opensurfaces}. The LMD comprises 5,845 low-resolution images acquired from indoor and outdoor sources, which have been manually labelled with 16 mutually exclusive material classes. On the other hand, OpenSurfaces contains 25,352 high-resolution indoor images labelled with 45 material categories. However, both datasets suffer from sparsely or coarsely labelled segments, as labelling images with material labels presents a significant challenge \cite{heng22camseg,schwartz2016material}. One of the key difficulties faced by annotators is that materials are often treated as properties of objects \cite{bell13opensurfaces,schwartz2016material}. Consequently, material segments tend to be labelled within object boundaries, which is undesirable as the material region should be marked independently of its context. For instance, when annotating a scene depicting a wooden bed on a wooden floor, the wood segment should ignore the object boundary and cover all wood pixels. Additionally, OpenSurfaces is highly unbalanced, with only 27 out of the 45 material classes having more than 60 samples. Among all the samples, 39.44\% are segmented as \textquotesingle wood\textquotesingle \ or \textquotesingle painted\textquotesingle. These limitations make the evaluation on OpenSurfaces less reliable compared to that on LMD. Therefore, this study mainly focuses on LMD, and the evaluation on OpenSurfaces will be presented as an additional piece of evidence. Notably, a recent dataset, MCubeS \cite{liang2022multimodal}, has been proposed to perform material segmentation on outdoor images using multimodal data such as imaging with near-infrared and polarised light. However, since this study concentrates on material segmentation using indoor RGB images, the evaluation on MCubeS is not included.

\subsection{Evaluation metrics} 
In this paper, the networks are evaluated with three metrics: mean pixel accuracy (Pixel Acc), mean class accuracy (Mean Acc), and mIoU. As discussed in Section \ref{dataset_dis}, the material annotations may not cover the whole material region. As a consequence, the mIoU numerator would be much smaller than it should be. This situation is especially severe for LMD which was annotated sparsely on purpose \cite{schwartz2016material,8675400}. Therefore, the mIoU is not reported for LMD. In addition to evaluating segmentation performance, we report the resources required for each model, including the number of trainable parameters and the number of floating-point operations (FLOPs) per forward propagation. To select SOTA models for evaluation, this paper sets a selection criterion with the frames per second (FPS). The model variants that can support real-time inference (FPS larger than 24) are compared with the DBAT in this paper.

\subsection{Implementation details} 
The networks reported in Section \ref{performance_analysis} are pre-trained on ImageNet \cite{deng2009imagenet}. This pre-training step is expected to teach the network with prior knowledge about contextual information such as scenes and objects. According to \cite{schwartz2016material}, contextual information can reduce the uncertainty in material segmentation. Therefore, the network should learn material features more efficiently with a pre-training strategy. The details will be discussed in Section \ref{netdissect_analysis}. For the Swin backbone, its implementation follows the original paper and the window size of the self-attention is set to 7. The decoder used in this paper is the Feature Pyramid Network (FPN) \cite{lin2017feature} for the reason that the FPN can recognise small material regions well \cite{heng22camseg}. The training images are resized first so that  the minimum borders are equal to 512. The resized images are then cropped into $512\times512$ patches to use batch training \cite{2014Efficient}. We use the AdamW optimiser to train the networks with batch size 16, coefficients $\beta_1$ 0.9, $\beta_2$ 0.999, and weight decay coefficient 0.01. Further, the learning rate is warmed-up from 0 to 0.00006 with 1,500 training steps, and decreased polynomially. The networks are trained on LMD for 200 epochs and on OpenSurfaces for five days.  


\section{Segmentation Performance Analysis}
\label{performance_analysis}
\subsection{Quantitative Analysis} 
Table \ref{testacc} reports the segmentation performance of our DBAT as well as five other models, ResNet-152 \cite{he2016deep}, ResNest-101 \cite{zhang2020resnest}, EfficientNet-b5 \cite{tan2019efficientnet}, Swin-t \cite{liu2021swin}, and CAM-SegNet \cite{heng22camseg}. Their heaviest variants that can serve real-time inference are evaluated apart from the CAM-SegNet. Although the CAM-SegNet does not meet the real-time selection criterion, it is the most recent architecture for the RGB-based material segmentation task. We notice that its architecture is suitable for our DBA module since it has a dedicated local branch. Therefore, in this paper, its local branch is equipped with the DBA module and its performance is reported as the CAM-SegNet-DBA \footnote{The CAM-SegNet-DBA is implemented by replacing the original local branch \cite{heng22camseg} by a combination of non-overlapping convolutional kernels and MLP. The patch resolution is enlarged by concatenating features within the kernel size.}. The evaluations are reported across five independent runs. The metric differences are reported in the order (Pixel Acc/Mean Acc/mIoU) with the additive method. 

As shown in Table \ref{testacc}, our DBAT achieves the best accuracy on the LMD among all the real-time models in terms of Pixel Acc and Mean Acc. Specifically, our DBAT achieves +0.85\%/+1.50\% higher than the second-best model, CAM-SegNet-DBA. It is also +2.54\%/+2.52\% higher than its backbone encoder, Swin-t. As for the OpenSurfaces, our DBAT beats the chosen models on Pixel Acc and mIoU. Moreover, its performance is comparable to the multi-branch CAM-SegNet-DBA (+0.50\%/-0.62\%/+0.09\%) with 9.65 more FPS and 19.6G fewer FLOPs. It is worth noting that compared with the performance reported in the original paper of CAM-SegNet \cite{heng22camseg}, our DBAT improves the Pixel Acc by 21.21\%. Moreover, the per-category analysis in Table \ref{append_table_per_category} shows that the DBA module improves the recognition of materials that usually have uniform appearances but varying shapes, such as paper, stone, fabric and wood. This indicates that the cross-resolution features successfully learn from distinguishable material features. 

\begin{figure*}
\centering
\noindent\resizebox{2\columnwidth}{!}{

\tikzset{every picture/.style={line width=0.75pt}} 

\begin{tikzpicture}[x=0.75pt,y=0.75pt,yscale=-1,xscale=1]

\draw (299.2,234.4) node  {\includegraphics[width=655.2pt,height=547.5pt]{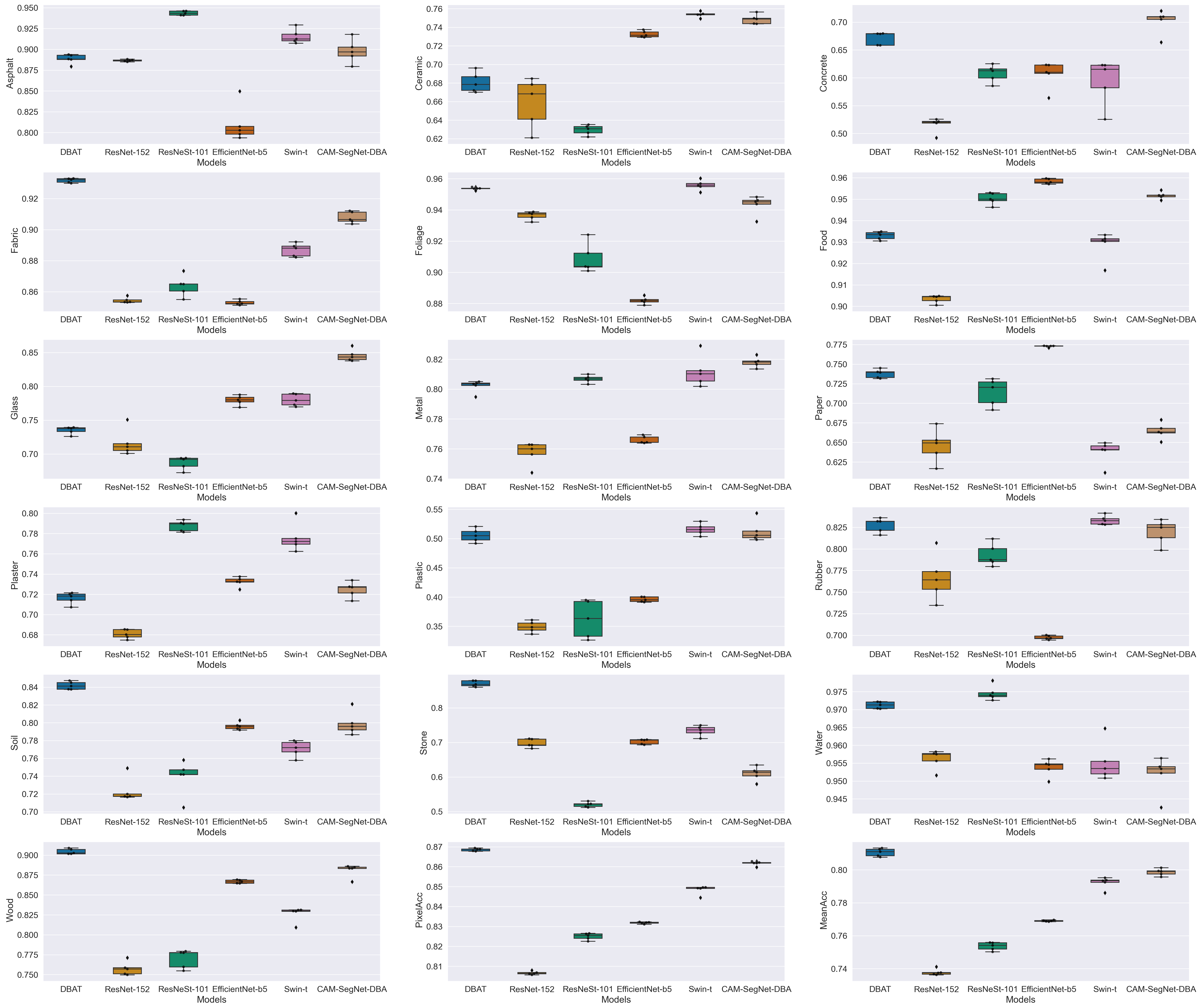}};
\end{tikzpicture}

}
\caption{Boxplot of the performance on  the LMD across five runs.}
\label{uncertainty_lmd}
\end{figure*}

\begin{figure*}
\centering
\noindent\resizebox{2\columnwidth}{!}{

\tikzset{every picture/.style={line width=0.75pt}} 

\begin{tikzpicture}[x=0.75pt,y=0.75pt,yscale=-1,xscale=1]

\draw (299.2,234.4) node  {\includegraphics[width=655.2pt,height=547.5pt]{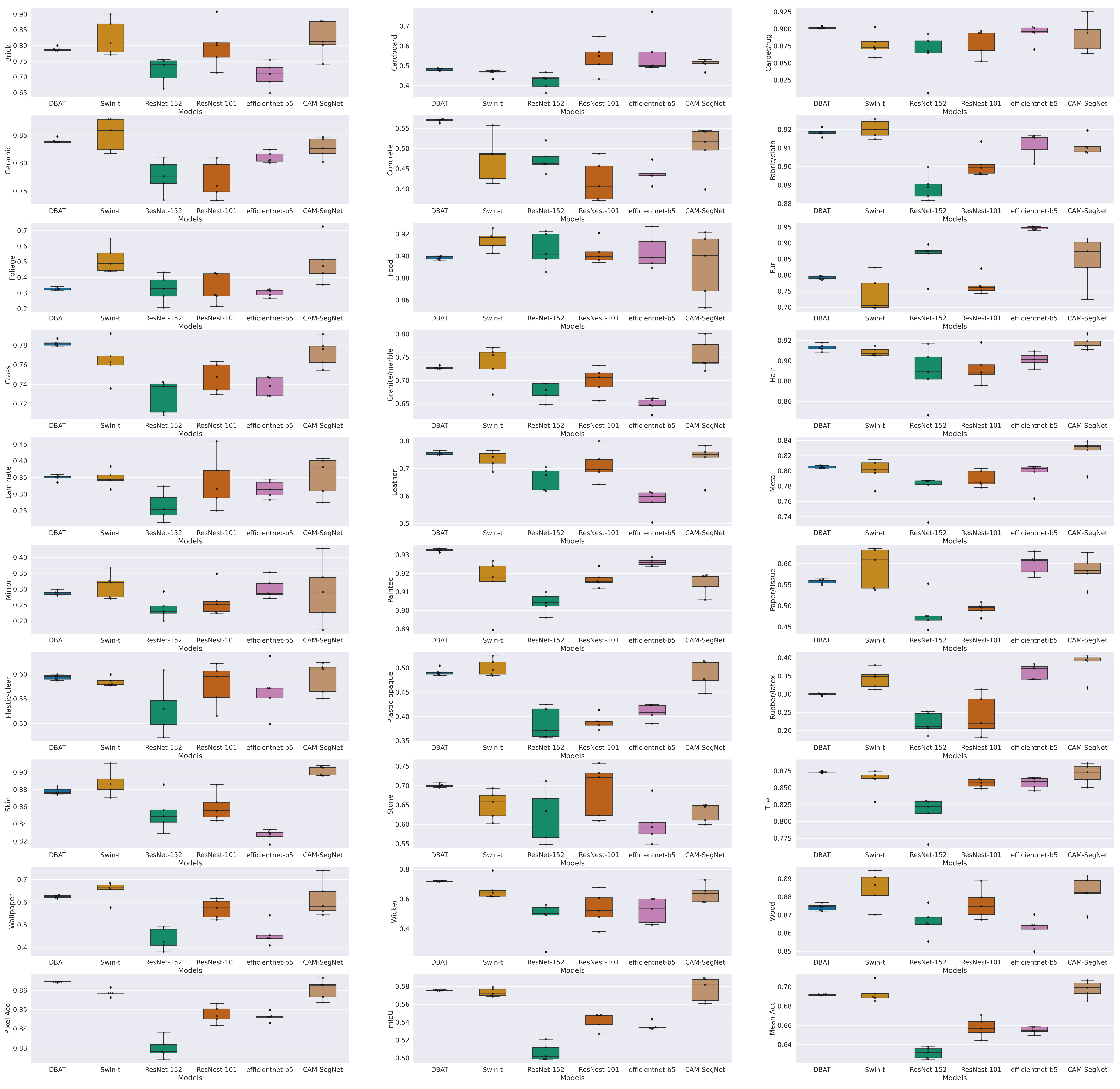}};
\end{tikzpicture}

}
\caption{Boxplot of the performance on  the OpenSurfaces across five runs.}
\label{uncertainty_opensurfaces}
\end{figure*}

As shown in Fig. \ref{uncertainty_lmd}, our DBAT can segment the materials consistently to achieve narrow uncertainty bounds, especially for the category foliage from LMD. Unlike the other five models, almost all the reported runs of our DBAT are within the upper and lower whiskers except for the category asphalt and metal. For the evaluations on OpenSurfaces shown in Fig. \ref{uncertainty_opensurfaces}, the uncertainty bounds of the proposed DBAT are much narrower compared with other models ($\pm$\SI{0.02}{\pp}  for Pixel Acc, $\pm$\SI{0.08}{\pp} for Mean Acc and $\pm$ \SI{0.06}{\pp} for mIoU)\footnote{Here p.p. stands for the percentage points.}. This indicates that the DBAT is robust to the network initialisation and can learn from image patches effectively. The CKA similarity score, which is 0.9583 for our DBAT, is calculated and averaged for every two checkpoints of the five individually trained networks to support this deduction.  

\begin{table*}
  \centering
\resizebox{1.8\columnwidth}{!}{%
\begin{tabular}{ c|cc|ccc|ccc} 
  \toprule
  Datasets & \multicolumn{2}{c|}{LMD}  & \multicolumn{3}{c|}{OpenSurfaces } & \multicolumn{3}{c}{--}  \\
  Architecture & Pixel Acc (\%) & Mean Acc (\%) & Pixel Acc (\%) & Mean Acc (\%) & mIoU (\%)	& \#params (M) & \#flops (G) & FPS \\
   \midrule 
ResNet-152        &  80.68 $\pm$ 0.11         & 73.87 $\pm$ 0.25  & 83.11 $\pm$ 0.68 & 63.13 $\pm$ 0.65 & 50.98 $\pm$ 1.12  & 60.75 & 70.27 & 31.35\\
ResNeSt-101        & 82.45 $\pm$ 0.20    & 75.31 $\pm$ 0.29          & 84.75 $\pm$ 0.57 & 65.76 $\pm$ 1.32 & 53.74 $\pm$ 1.06 & 48.84 & 63.39 &25.57\\
EfficientNet-b5    &   83.17 $\pm$ 0.06        &  76.91 $\pm$ 0.06         &  84.64 $\pm$ 0.34 & 65.41 $\pm$ 0.44   & 53.79 $\pm$ 0.54  &  30.17 & 20.5 &27.00\\

Swin-t             &  84.70 $\pm$ 0.26       & 79.06 $\pm$ 0.46         & 85.88 $\pm$ 0.27 & \textbf{69.74 $\pm$ 1.19} &  57.39 $\pm$ 0.54     &29.52 & 34.25&33.94\\
CAM-SegNet-DBA & \underline{86.12 $\pm$ 0.15} & \underline{79.85 $\pm$ 0.28} & \underline{86.00 $\pm$ 0.64} & \underline{69.61 $\pm$ 1.08}  & \underline{57.52 $\pm$ 1.44} & 68.58 & 60.83 & 17.79\\
DBAT           & \textbf{86.85 $\pm$ 0.08} & \textbf{81.05 $\pm$ 0.28} & \textbf{86.43 $\pm$ 0.02} & 69.18 $\pm$ 0.08 &  \textbf{57.57 $\pm$ 0.06}  & 56.03 & 41.23&27.44\\

 \bottomrule
 \end{tabular}
 }
  \vspace{3pt}
  \caption{Segmentation performance on the LMD and the OpenSurfaces. The FPS is calculated by processing 1000 images with one NVIDIA 3060ti. The uncertainty evaluation is reported by training the networks five times. The best performance is shown in bold text and the second best is underlined.}
  \label{testacc}
\end{table*}

\begin{table*}
  \centering
\resizebox{1.8\columnwidth}{!}{%
    \begin{tabular}{c|cccccc}
    \toprule
        Model  & ResNet-152 & ResNeSt-101 & EfficientNet-B5 & Swin-t & CAM-SegNet-DBA & DBAT \\ 
        \midrule
        Asphalt  & 88.66 $\pm$ 0.17 & \textbf{94.35 $\pm$ 0.27} & 82.17 $\pm$ 2.80 & \underline{91.83 $\pm$ 1.09} & 89.87 $\pm$ 1.94 & 88.66 $\pm$ 0.72 \\ 
        Ceramic  & 65.29 $\pm$ 3.19 & 62.86 $\pm$ 0.67 & 73.34 $\pm$ 0.42 & \textbf{75.35 $\pm$ 0.42} & \underline{75.01 $\pm$ 0.64} & 68.31 $\pm$ 1.31 \\ 
        Concrete  & 50.89 $\pm$ 1.67 & 60.53 $\pm$ 2.00 & 59.36 $\pm$ 2.98 & 57.42 $\pm$ 4.88 &  \textbf{69.20 $\pm$ 2.81} & \underline{66.90 $\pm$ 1.07} \\ 
        Fabric  & 85.53 $\pm$ 0.22 & 86.420 $\pm$ 0.92 & 85.33 $\pm$ 0.20 & 88.71 $\pm$ 0.50 & \underline{90.79 $\pm$ 0.43} & \textbf{93.14 $\pm$ 0.16}\\ 
        Foliage & 93.55 $\pm$ 0.33 & 91.25 $\pm$ 1.16 & 88.21 $\pm$ 0.32 & \textbf{95.57 $\pm$ 0.45} & 94.04 $\pm$ 0.79 & \underline{95.35 $\pm$ 0.12}  \\ 
        Food  & 90.27 $\pm$ 0.22 & 94.96 $\pm$ 0.34 & \textbf{95.84 $\pm$ 0.14} & 92.51 $\pm$ 0.83 & \underline{95.19 $\pm$ 0.24} & 93.27 $\pm$ 0.22 \\ 
        Glass & 72.58 $\pm$ 2.50 & 68.33 $\pm$ 0.34 & 77.83 $\pm$ 0.94 & \underline{77.95 $\pm$ 0.99} & \textbf{84.88 $\pm$ 1.11} & 73.27 $\pm$ 0.67  \\ 
        Metal & 75.35 $\pm$ 0.94 & 80.66 $\pm$ 0.34 & 76.67 $\pm$ 0.28 & \underline{81.54 $\pm$ 1.36} & \textbf{81.83 $\pm$ 0.48} & 79.99 $\pm$ 0.51  \\ 
        Paper  & 64.52 $\pm$ 2.87 & 71.14 $\pm$ 1.99 & \textbf{77.21 $\pm$ 0.13} & 63.05 $\pm$ 1.90 & 66.48 $\pm$ 1.43 & \underline{73.83 $\pm$ 0.67} \\ 
        Plaster  & 68.01 $\pm$ 0.53 & \textbf{78.76 $\pm$ 0.62} & 73.11 $\pm$ 0.64 & \underline{78.12 $\pm$ 1.90} & 72.37 $\pm$ 1.03 & 71.43 $\pm$ 0.71 \\ 
        Plastic  & 34.87 $\pm$ 1.21 & 36.07 $\pm$ 3.42 & 39.59 $\pm$ 0.64 & \underline{51.64 $\pm$ 1.31} & \textbf{52.07 $\pm$ 2.28} & 50.62 $\pm$ 1.45 \\ 
        Rubber  & 77.08 $\pm$ 3.61 & 79.57 $\pm$ 1.62 & 69.73 $\pm$ 0.29 & \textbf{83.48 $\pm$ 0.67} & 81.63 $\pm$ 1.79 & \underline{82.61 $\pm$ 1.01} \\ 
        Soil  & 73.27 $\pm$ 1.63 & 73.15 $\pm$ 2.67 & 79.73 $\pm$ 0.55 & 76.89 $\pm$ 1.11 & \underline{80.39 $\pm$ 1.73} & \textbf{84.25 $\pm$ 0.50} \\ 
        Stone  & 69.66 $\pm$ 1.42 & 52.12 $\pm$ 0.93 & 70.07 $\pm$ 0.76 & \underline{73.05 $\pm$ 1.92} & 60.73 $\pm$ 2.76 & \textbf{86.94 $\pm$ 0.95} \\ 
        Water  & 95.49 $\pm$ 0.33 & \textbf{97.54 $\pm$ 0.28} & 95.30 $\pm$ 0.32 & 95.78 $\pm$ 0.70 & 94.95 $\pm$ 0.69 & \underline{97.12 $\pm$ 0.10} \\ 
        Wood  & 76.05 $\pm$ 1.08 & 76.71 $\pm$ 1.23 & 86.69 $\pm$ 0.24 & 82.03 $\pm$ 1.11 & \underline{87.63 $\pm$ 0.98} & \textbf{90.53 $\pm$ 0.37} \\ 
        \midrule
        Pixel Acc & 80.68 $\pm$ 0.11 & 82.45 $\pm$ 0.20 & 83.17 $\pm$ 0.06 & 84.71 $\pm$ 0.26 & \underline{86.12 $\pm$ 0.15}  & \textbf{86.85 $\pm$ 0.08} \\ 
        Mean Acc & 73.87 $\pm$ 0.25 & 75.31 $\pm$ 0.29 & 76.91 $\pm$ 0.06 & 79.06 $\pm$ 0.46 & \underline{79.85 $\pm$ 0.28} & \textbf{81.05 $\pm$ 0.28}  \\ 
        \bottomrule
    \end{tabular}
    }
    \vspace{3pt}
    \caption{Per-category performance analysis in terms of Pixel Acc (\%). The networks are trained five times to report the uncertainty. The metrics are reported in percentages.}
    \label{append_table_per_category}
\end{table*}

\subsection{Qualitative Analysis} 
Fig. \ref{B-4.1} shows the predicted material segmentation for three images. In Fig. \ref{B-4.1} (a), ResNet-152 \cite{he2016deep}, ResNest-101 \cite{zhang2020resnest}, Swin-t \cite{liu2021swin}, and the modified CAM-SegNet-DBA segment the bed as fabric, the floor as plaster. However, in this image, the floor appears to be covered with a carpet whose material is fabric. One possible explanation is that for the scene bedroom, the floor and wall are typically covered with plaster. These networks fail to make predictions based on material features, but rely on contextual information, so they tend to use plaster as a label for predictions. The proposed DBAT and the EfficientNet-b5 \cite{tan2019efficientnet} break the object boundary and segment part of the floor as fabric. Moreover, there are fewer noisy pixels in the DBAT segmented image when compared with the EfficientNet-b5. This indicates that with cross-resolution patch features, the DBAT can identify materials densely and precisely even if it is trained on a sparsely labelled dataset.

The segmented materials in Fig. \ref{B-4.1} (b) and Fig. \ref{B-4.1} (c) provide more evidence that the DBAT can segment the images well with features extracted from cross-resolution patches. In Fig. \ref{B-4.1} (b), the boundary between the wooden window frame and glass-made windows in DBAT segmented image is more adequate than the segments predicted by other networks. In Fig. \ref{B-4.1} (c), the segmented fabric aligns well with the ground truth with no noisy predictions. Considering that the training of DBAT takes sparsely labelled segments, it is reasonable to say that DBAT learns the difference between materials from cross-solution patches.

\subimport{../tikzfiles/}{qualitative_eva.tex}

\subsection{Ablation Study} 
This section studies the effectiveness of each component of the DBAT. In Table \ref{ablationacc}, the performance is reported after removing the feature merging module and the DBA module in sequence. Without the feature merging module, the Pixel Acc  and Mean Acc  decrease by \SI{1.61}{\pp} and \SI{2.04}{\pp} respectively. This shows the importance of the attention-based residual connection in improving performance. The performance drops by another \SI{0.72}{\pp} in Pixel Acc and \SI{0.13}{\pp} in Mean Acc after removing the DBA module. This shows that the DBA module that learns complementary cross-resolution features guided by the feature merging module can improve performance effectively.

\begin{table}[!htb]
  \centering
\begin{tabular}{ c|cc} 
  \toprule
  Architecture & $\Delta$ Pixel Acc & $\Delta$ Mean Acc \\
   \midrule 
- Feature merging       &  -1.61 & -2.04 \\
- Dynamic backward attention  &  -2.33        & -2.17   \\

 \bottomrule
 \end{tabular}
  \vspace{3pt}
  \caption{The ablation study to analyse each component of our DBAT. The performance difference is reported in percentage points.}
  \label{ablationacc}
\end{table}

In order to justify the network design, this section further studies the alternative implementations of the DBAT from three aspects: 1) how the attention masks are predicted; 2) how the feature maps are down-sampled; 3) how aggregated features are merged with $Map_4$. The performance differences in Table \ref{choices} are reported by switching one of the implementations in DBAT to its alternatives. As stated in Section \ref{dbasection}, the proposed DBAT adopts convolutional kernels to generate the per-pixel attention masks \cite{chen2020dynamic}. By replacing the kernels to dilated ones \cite{wei2018revisiting}, the receptive field is enlarged when predicting the masks. However, the DBAT performance decreases significantly by \SI{-2.15}{\pp}/\SI{-2.67}{\pp}. This indicates that the local information is critical for the dynamic attention module to work well. Section \ref{dbasection} describes that the cross-resolution feature maps need to be down-sampled so that the fixed-size attention masks can be applied. Originally DBAT uses the MLP to down-sample the feature map. Instead of using such a trainable method, we can also use a superficial non-parametric pooling layer, which decreases the performance by \SI{-0.88}{\pp}/\SI{-1.58}{\pp}. The slight drop in Pixel Acc and the significant drop in Mean Acc suggest that the trainable down-sampling method can help balance the performance of different material categories. As for the feature merging module, a simple residual connection slightly reduces the performance by \SI{-0.58}{\pp}/\SI{-0.64}{\pp}. This highlights that DBAT needs a residual connection to guide the aggregation of cross-resolution features and the representation capability of the feature merging module may not be vital.

\begin{table}[!htb]
  \centering
\resizebox{\columnwidth}{!}{%
\begin{tabular}{ c|c|cc} 
  \toprule
  \multicolumn{2}{c|}{Implementation Choices}  & $\Delta$ Pixel Acc & $\Delta$ Mean Acc  \\
   \midrule 

\begin{tabular}{c}Generate\\Attention Masks\end{tabular}   & CNN $-->$  Dilated  CNN      &    -2.15   & -2.67 \\
  
  \midrule 
Down-sample     & MLP $-->$ Average Pooling  & -0.88 & -1.58  \\

\midrule 
Feature Merging    & Attention $-->$ Residual Connection   &    -0.58   &  -0.64\\
 \bottomrule
 \end{tabular}
  }
  \vspace{3pt}
  \caption{The study of implementation choices in each component of our DBAT. The performance difference is reported in percentage points.}
  \label{choices}
\end{table}

\section{Network Behaviour Analysis}
Although our DBAT achieves the best accuracy with carefully designed propagation, like other successful networks, its mechanism is still hidden in a black box. In order to interpret the DBAT, we employ both statistical methods such as reporting the average attention weights for each patch resolution and feature interpretability methods such as the CKA analysis and network dissection. We discover that our DBAT is particularly good at learning material-related features such as texture, which justifies our assumption that features extracted from image patches can boost performance in the material segmentation task.
\subsection{CKA Heatmap} 
Fig. \ref{fig:ckamatrix} (a) visualises the CKA heatmap of the DBAT and Fig. \ref{fig:ckamatrix} (b) shows the heatmap measured between DBAT and Swin. The brightness of the colour indicates the similarity of the features extracted by two layers. The network layers are indexed by the forward-propagation order. The DBAT and the Swin share the same network architecture before layer 106, which explains the  bright diagonal in Fig. \ref{fig:ckamatrix}  (b) connecting (0, 0) and (105, 105). When observed closely, the diagonal line becomes darker as it approaches the point (105, 105). This suggests that at shallower layers, the Swin backbone extracts features similar to when it is used alone, and it gradually learns something new as it approaches the deeper layers. The dark areas from layer 106 to layer 113 in Fig. \ref{fig:ckamatrix} (a) reflect the attention masks predicted by the DBA module. By collecting the cross-resolution feature maps, the aggregated features contain information from both shallow and deep layers, illustrated by the bright region between layer 113 and 124. After layer 124, the feature merging module injects the relevant information from the aggregated features into $Map_4$. This module produces a feature map that differs from the features extracted by Swin, as shown by the darker points near (140, 100) in Fig. \ref{fig:ckamatrix} (b). 
\subimport{../tikzfiles/}{ckaheatmap.tex}
\subsection{Attention Analysis} 
\label{attention_analysis}
This section analyses the dynamic attention module by visualising the attention masks and calculating their descriptive statistics, including the average attention weights as well as the equivalent attention patch resolutions. Fig. \ref{fig:attentionmasks} shows the attention masks for images in the LMD test set. The patch resolution increases from $Attn_1$ to $Attn_4$. We notice that a prediction for a material covering multiple small objects or a small area tends to depend on the features extracted from small patch resolutions. For example, the first column images in Fig. \ref{fig:attentionmasks} highlight the regions on where the network concentrates. A brighter colour indicates higher attention weights. The wooden area in the first row covers both the desk and the floor. The wooden chairs and the fabric floor are mutually overlapping in the second row.  Small patches can isolate the objects and learn material features along the boundaries.

\subimport{../tikzfiles/}{attention_vis.tex}

Fig. \ref{fig:analysis_attn} shows  the average attention weight in (a) and the equivalent patch size in (b, the box plot). The equivalent patch size, which represents the attention distance of each attention head, is calculated by transforming the attention diagonal distance in \cite{raghu2021vision} to the side length of a square, which suggests the patch resolution used to extract features from the feature maps $Map_{1,2,3,4}$. As expected, the aggregated features mostly (52.40\%) depend on $Attn_4$, which is thoroughly processed by the whole backbone encoder, with an average patch size of 74.31. Apart from $Attn_4$, the aggregated features also depend on feature maps extracted from small patch sizes. For example, $Attn_3$ is extracted from an average patch size of 31.68, and it contributes 30.50\% to the aggregated features. Although $Attn_1$ is gathered from a shallow stage of the network, the aggregated features still depend on it to handle the overlapping material regions with a patch size of 6.75 on average. 

To further illustrate the effect of the dynamic backward attention module, the similarity scores comparing one layer with $Map_{1,2,3,4}$ from the CKA matrix is reported in Fig. \ref{fig:analysis_attn}. (b, the line plot). The blue line compares $Map_4$ from the Swin encoder, and the orange line compares the aggregated features from the DBAT. The increased similarity scores against $Map_{1,2,3}$ clearly show that the aggregated feature depends on information from shallow layers.

\subimport{../tikzfiles/}{attention_descriptive.tex}

\subsection{Network Dissection} 
\label{netdissect_analysis}
The network dissection method aligns the disentangled neurons of one network layer to semantic concepts \cite{bau2017network,bau2019gan}. By counting the portion of neurons aligned to each concept, it is possible to give an understanding of what features the network learns. This paper studies local concepts such as colour, texture and part, as well as global/semi-global concepts like object and scene. The neurons of the last encoder layer are selected to be analysed since they are more interpretable than shallow layers \cite{bau2017network,bau2019gan}. Fig. \ref{netdissec-swindbat} (a) depicts how pre-training influences the features that the DBAT learns. Without pre-training, DBAT (the blue line in Fig. \ref{netdissec-swindbat}) has shown a preference for texture features, and a portion of its neurons can detect object and scene features. As shown by the dotted orange line, the Swin backbone \cite{liu2021swin} trained with ImageNet \cite{martinez2017image} tends to detect object features. The DBAT has more neurons aligned to texture and object features when trained with a pre-trained backbone, shown as the purple line. The observations indicate two aspects: (1) the DBAT relies on texture features to solve the material segmentation task. (2) the pre-trained object detectors reduce the uncertainty in identifying materials and ease the training of texture detectors.

This paper further analyses the feature difference between material and object tasks. In particular, Swin \cite{liu2021swin} is trained with two object-related datasets: the ImageNet \cite{deng2009imagenet}, an object classification dataset, and the ADE20K \cite{zhou2017scene,zhou2019semantic}, an object-level segmentation dataset. As shown in Fig. \ref{netdissec-swindbat} (b), the significant difference is the lack of texture features in object-related tasks. This discovery highlights that enhancing features hidden in patches is a valid heuristic to improve the network performance on the material segmentation task. 

Fig. \ref{netdissec-comparemodels} dissects three more networks: ResNet-152 \cite{he2016deep}, ResNest-101 \cite{zhang2020resnest}, EfficientNet-b5 \cite{tan2019efficientnet} on both material and object tasks. An interesting discovery is that although these networks achieve comparable performance, the features that they learn are different. For example, the ResNet-152 relies on texture features on both tasks and it learns more part-related features on material task compared with object task. Although three of the networks in Fig. \ref{netdissec-comparemodels} learn texture features on material task, a special exception is the EfficientNet-b5, which knows almost nothing about texture for both tasks. This phenomenon goes against the intuition that networks targeting material segmentation should learn texture features well since texture describes the appearances of materials. One reasonable explanation is that assigning material labels to object or object parts can cover the labelled material region and achieves a high accuracy since these material datasets are sparsely labelled. Therefore, here we call for densely labelled material segmentation datasets for reliable evaluation and analysis.

\subimport{../tikzfiles/}{netdissec_swindbat.tex}

\subimport{../tikzfiles/}{netdissec_comparemodels.tex}




\section{Conclusion}
This paper proposed a single-branch network, the DBAT, to learn material-related features extracted from image patches at multiple resolutions. The features are aggregated dynamically based on predicted attention weights. The network is designed to learn complementary features from the cross-resolution patches with an attention-based residual connection. The DBAT outperforms all chosen real-time models evaluated on two datasets. It also achieves comparable performance with fewer FLOPs compared with the multi-branch network, the CAM-SegNet-DBA. As for network analysis, this paper illustrates its inference pipeline by visualising the attention masks as well as the CKA heatmaps. The experiment shows that about half of the information comes from small image patches, and those cross-resolution features can help the network learns from overlapping materials. Moreover, the CKA heatmap shows that the aggregated features carry the knowledge from shallow layers, which may be the key to achieving good performance on material segmentation tasks. Instead of focusing solely on achieving better performance on material segmentation tasks, we moved a step further in this paper to interpret the material features that a network learns with semantic concepts. The counted portions of neurons aligned to each concept show that our DABT is particularly good at extracting texture-related features. The analysis also reflects the potential unreliability of evaluating sparse datasets, so we will seek for densely labelled material datasets for reliable network evaluations in future work. One possible method is to use the physically-based rendering to generate synthetic densely annotated material segmentation datasets.

\section*{Acknowledgments}
This work was supported by the EPSRC Programme Grant Immersive Audio-Visual 3D Scene Reproduction Using a Single 360 Camera (EP/V03538X/1). For the purpose of open access, the author has applied a Creative Commons Attribution* (CC BY) licence to any Author Accepted Manuscript version arising.

\bibliography{egbib}
\bibliographystyle{IEEEtran}

\vfill

\end{document}